\documentclass[letterpaper, 10 pt, conference]{ieeeconf}  

\IEEEoverridecommandlockouts                              
\UseRawInputEncoding
\usepackage{cite}

\usepackage{graphicx} 
\usepackage{subfigure}
\usepackage{amsmath} 
\usepackage{amssymb}  
\usepackage{booktabs}
\usepackage{multirow}

\title{\LARGE \bf
SQ-SLAM: Monocular Semantic SLAM Based on Superquadric Object Representation
}

\author{Xiao Han$^{}$ and Lu Yang$^{*}$
\thanks{*Corresponding author.}
\thanks{Authors are with the School of Automation Engineering, University of Electronic Science and Technology of China, Chengdu, China.} 
    }

\begin{document}

\maketitle
\thispagestyle{empty}
\pagestyle{empty}

\begin{abstract}
Object SLAM uses additional semantic information to detect and map objects in the scene, in order to improve the system's perception and map representation capabilities. Quadrics and cubes are often used to represent objects, but their single shape limits the accuracy of object map and thus affects the application of downstream tasks. In this paper, we introduce superquadrics (SQ) with shape parameters into SLAM for representing objects, and propose a separate parameter estimation method that can accurately estimate object pose and adapt to different shapes. Furthermore, we present a lightweight data association strategy for correctly associating semantic observations in multiple views with object landmarks. We implement a monocular semantic SLAM system with real-time performance and conduct comprehensive experiments on public datasets. The results show that our method is able to build accurate object map and has advantages in object representation. Code will be released upon acceptance.

\end{abstract}

\section{INTRODUCTION}

Visual simultaneous localization and mapping (SLAM) is one of the basic technologies of robot spatial perception, which has been widely used in mobile robots, autonomous driving and AR/VR. 
Traditional methods use geometric primitives such as points \cite{orb}, lines \cite{pumarola2017pl} and planes \cite{yunus2021manhattanslam} in the scene to build maps and simultaneously estimate the pose of sensors. Its sparse, semi-dense or dense metric maps focus on building an accurate geometric representation of the world.
However, maps containing only pure geometric information limit high-level tasks. For example, mobile robots need to perceive semantic information for task-oriented navigation.
The rapid development of deep learning in recent years has laid the foundation for further enhancing the robot's perception and map representation capabilities. The robotics community is increasingly introducing learning-based methods \cite{redmon2016you,he2017mask} into SLAM systems.

Object SLAM is one of the competitive applications that combine semantic information with visual SLAM.
Objects are introduced into the observation process as map elements, which improves the robot's understanding of the scene and is helpful to perform more complex tasks \cite{wu2021object}. On the other hand, the observation of some long-term consistent objects has a positive effect on the long-term operation and relocalization of SLAM system \cite{qian2022towards}.
In recent work, objects are represented by cubes \cite{yang2019cubeslam,wu2020eao} or ellipsoids \cite{hosseinzadeh2019real,ok2019robust,nicholson2018quadricslam}, and object map is constructed by only 2D bounding box without prior 3D models. This compact geometric representation retains the basic space occupation information of objects. However, simple cubes or ellipsoids cannot effectively represent the shapes of various objects, which may lead to inaccurate scene understanding or affect object interaction.

In this work, we propose a monocular semantic SLAM based on superquadric object representation. Superquadrics are extensions to standard quadrics, which can represent common 3D geometric primitives such as ellipsoids, cylinders and cubes. We introduce it into object SLAM to accurately represent different kinds of objects with a set of abstract geometric parameters.
The parameters are estimated separately in the SLAM front-end and back-end threads to effectively implement superquadrics based on optimization and ensure real-time performance. In addition, data association is another challenge in object SLAM. We propose a lightweight data association strategy that includes inter-frame association using object appearance information and non inter-frame association using object geometry information.
Comprehensive experiments on public datasets demonstrate that our method outperforms baselines in terms of object representation strength.

The contributions of this work are as follows:
\begin{itemize}
	
\item Propose a superquadric object representation method with separate parameter estimation, which can construct semantic enhanced object map.
\item We leverage the appearance information of objects and the geometric information of sparse point cloud to implement a lightweight data association strategy.
\item Based on the proposed algorithm, we implement a real-time monocular object SLAM, and verify its effectiveness on public datasets.

\end{itemize}

\section{RELATED WORK}

\subsection{Object Representation in SLAM}

There are different representations of objects in many literatures on semantic SLAM. 
Salas et al. \cite{salas2013slam++} first introduced objects as landmarks into SLAM system, which uses RGB-D information and prior object models to build maps, but it cannot be applied to general scenes.
Maskfusion \cite{runz2018maskfusion} accurately constructs dense object models with the help of semantic segmentation, which are represented as surfel clouds.
In contrast, Sunderhauf et al. \cite{sunderhauf2017meaningful} avoid pixel-by-pixel processing and perform unsupervised segmentation only on dense point clouds located in detection bounding boxes. Point cloud clusters with semantic labels are used to represent objects, but they do not have spatial metric attributes.

Implementing object SLAM using only monocular camera has also attracted many interests. In 2019, Yang et al. Proposed CubeSLAM \cite{yang2019cubeslam}, which uses 3D cubes inferred from 2D bounding boxes as object representation. Compared with dense models, this simple geometric primitive retains the basic position and size information of objects. In addition, quadrics are favored by researchers due to their simple mathematical representation and compact perspective projection model. QuadricSLAM \cite{nicholson2018quadricslam} introduces quadrics into SLAM as object landmarks, and optimizes their parameters through continuous image observation after initialization. Recently, researchers have focused on improving the accuracy and robustness of quadric landmarks. Tian et al. \cite{TianAccurate} proposed a parameter-separated initialization method to make it suitable for vehicles in outdoor environments. SO-SLAM \cite{liao2022so} explores the symmetry of objects and implements an orientation fine-tuning algorithm. Hu et al. \cite{hu2022making} introduced symmetric positive-definite matrix manifold to solve the singularity of classical quadric parameterization method. 

However, the single geometric model used in the above methods is difficult to fit various types of objects. Zhen et al. \cite{zhen2022unified} used quadrics and their degenerate cases to uniformly represent geometric primitives like planes, ellipsoids, and cylinders. Tschopp et al. \cite{tschopp2021superquadric} explored superquadrics but did not build a complete object SLAM system.

\subsection{Data Association}

The goal of data association is to correctly associate multiple object observations at different positions and times with the same object landmark. Bowman et al. \cite{bowman2017probabilistic} Proposed an expectation maximization (EM) algorithm for soft data association, and tightly coupled metric information, semantic information and data association into a unified optimization framework. Doherty et al. \cite{doherty2020probabilistic} proposed the max-marginalization procedure and semantic max-mixture factors for optimization. However, these algorithms based on EM or max-marginalization need many computing resources.

Data associations can also be solved directly by object observations. \cite{qian2021semantic} calculates the Bag of Words (BoW) vector of feature points as the appearance description of objects, and then perform BoW matching on the candidate objects that satisfy the reprojection relationship. Chen et al. \cite{chen2022accurate} Proposed a hierarchical object association strategy, and applied multi-object tracking for short-term object association. Although these methods based on object appearance are efficient, they are sensitive to observation noise and not robust enough. Iqbal et al. \cite{iqbal2018localization} used nonparametric statistical test to address the non-Gaussian property of object point cloud distributions. \cite{wu2020eao} extends it, taking into account the statistical properties of the point cloud centroid. Inspired by \cite{wu2020eao}, this work leverages the appearance and geometric information of objects, and the implemented association algorithm is lightweight to ensure real-time performance.

\section{SUPERQUADRICS}

\begin{figure}[t]
	\centering
	\includegraphics[width=2.6in]{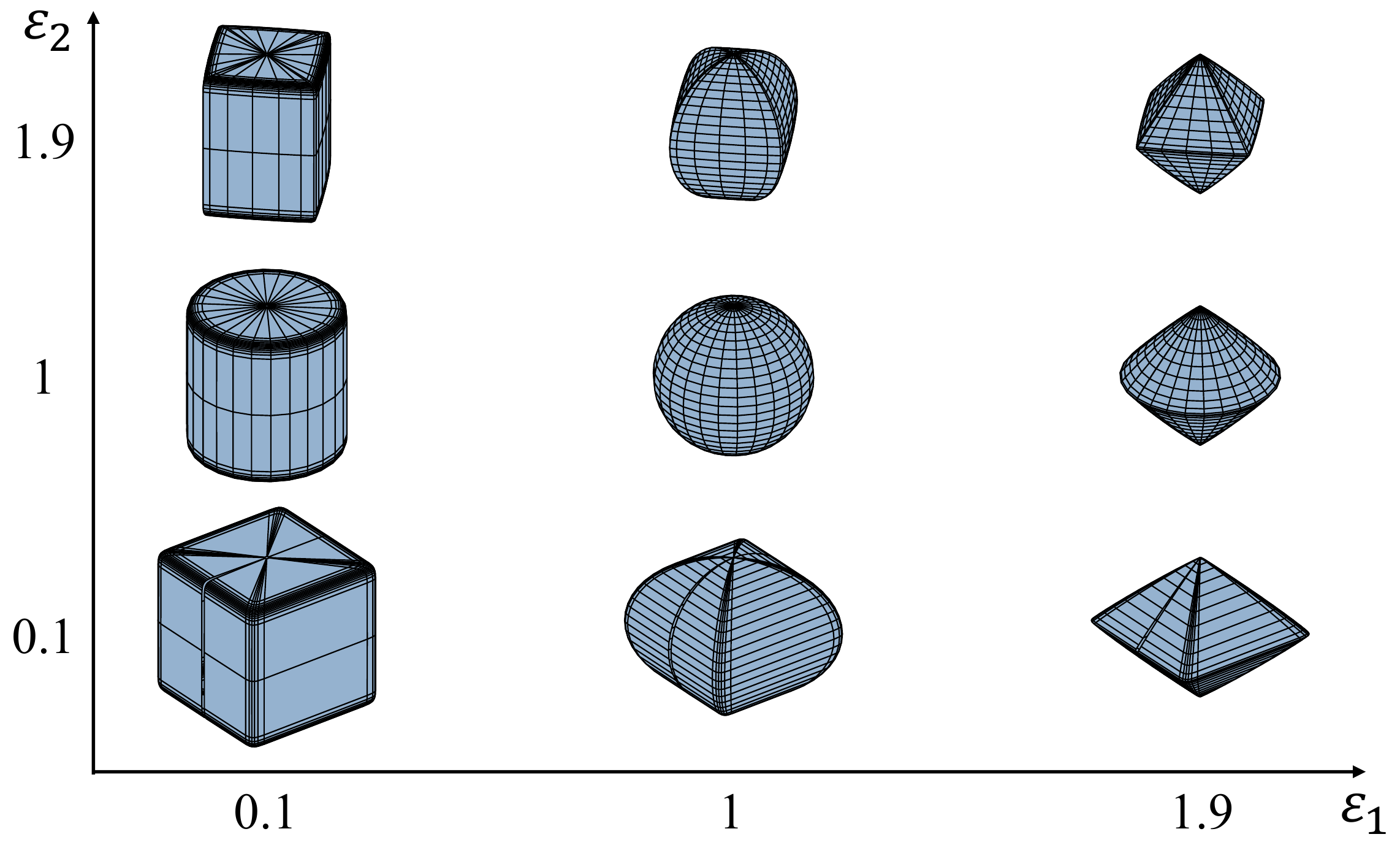}
	\caption{Superquadrics with different values of shape parameters $\varepsilon_{1}$ and $\varepsilon_{2}$. Size parameters $a_x$, $a_y$ and $a_z$ are equal.}
	\label{fig2}
\end{figure}

Superquadrics \cite{barr1981superquadrics} are a series of parametric surfaces that include superellipsoids, supertoroids, and superhyperboloids with one piece and two pieces. In this paper, we focus on superellipsoids, and follow community convention to refer to them by the more generic term of superquadrics.
A superquadric can be obtained by a spherical product of two superellipses, $\mathbf{s} _{1}$ and $\mathbf{s} _{2}$, which is described by the following parametric equation:
\begin{equation}
\mathbf{p}(\eta,\omega ) = \mathbf{s} _{1}(\eta) \otimes \mathbf{s} _{2}(\omega)=
\begin{bmatrix}a_{x}\cos^{\varepsilon _{1}}\eta
\cos^{\varepsilon _{2}}\omega \\ a_{y}\cos^{\varepsilon _{1}}\eta \sin^{\varepsilon _{2}}\omega
\\a_{z} \sin^{\varepsilon _{1}}\eta 
\end{bmatrix}
\end{equation}
where $\mathbf{p}(\eta,\omega )$ denotes a point on the superquadric surface determined by two angle variables ($-\frac{\pi }{2}  \le \eta\le \frac{\pi }{2}, -\pi   \le \omega \le \pi$). The parameters $a_{x}, a_{y}$ and $a_{z}$ define the size of the superquadric in three directions respectively, while $\varepsilon_{1} $ and $\varepsilon_{2}$ control the shape.
Since the shapes of most common objects are convex, and in order to ensure numerical stability during the optimization process \cite{vaskevicius2017revisiting}, we limit the value range of shape parameters, i.e. $  0.1\le \varepsilon_{1} ,\varepsilon_{2} \le1.9 $.
The shapes corresponding to different $\varepsilon_{1} $ and $\varepsilon_{2}$ are shown in Fig. \ref{fig2}. When $\varepsilon_{1}=\varepsilon_{2}=1$, it degenerates into a standard ellipsoid.
Superquadrics can also be expressed by the following implicit equation:
\begin{equation}
\mathit{F} (x,y,z) =  \left (   \left(\frac{x}{a_{x}} \right)^{\frac{2}{\varepsilon_{2}} } + 
\left(\frac{y}{a_{y}} \right)^{\frac{2}{\varepsilon_{2}} }     \right ) ^{\frac{\varepsilon_{2}}{\varepsilon_{1}} }
+\left(\frac{z}{a_{z}} \right)^{\frac{2}{\varepsilon_{1}} }  = 1
\label{eq2}
\end{equation}
This function is also called inside-outside function, because it can be used to determine where the point lies with respect to the superquadric surface.

Defining a superquadric in the world coordinate system requires determining its rigid motion relative to the origin, including $t_x,t_y,t_z$ for the translation and $ \rho, \psi, \theta$ for determining the rotation. A total of 11 parameters can represent a superquadric in general position as
\begin{equation}
\mathit{\Lambda } =\left \{a_x,a_y,a_z,\varepsilon_1,\varepsilon_2,\rho, \psi, \theta,t_x,t_y,t_z   \right \}.
\end{equation}
For more properties of superquadrics, please refer to \cite{jaklivc2000superquadrics}.

\section{SQ-SLAM}

SQ-SLAM is built on the feature point-based ORB-SLAM2, and uses bounding boxes output by object detector \cite{Redmon_2016_CVPR} as additional observations to simultaneously estimate camera pose and objects. The superquadric parameters are typically retrieved from depth information \cite{duncan2013multi,makhal2018grasping}. However, it is difficult to estimate all 11 parameters of superquadric directly from sparse point clouds. Considering the real-time requirement of SLAM, we estimate the superquadric parameters separately in two parallel threads. The tracking thread only estimates object pose to ensure accurate data association and efficiency. The local mapping thread jointly optimizes camera pose and object shape to achieve accurate object representation and build a semantically enhanced object map.

\begin{figure}[t]
	\centering
	\subfigure[]
	{
		\centering
		\includegraphics[width=1.55in]{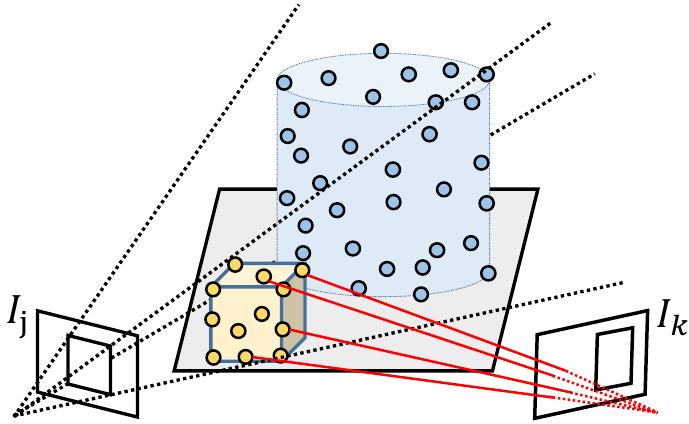}
		\label{fig3:a}
	}
	\subfigure[]
	{
		\centering
		\includegraphics[width=1.2in]{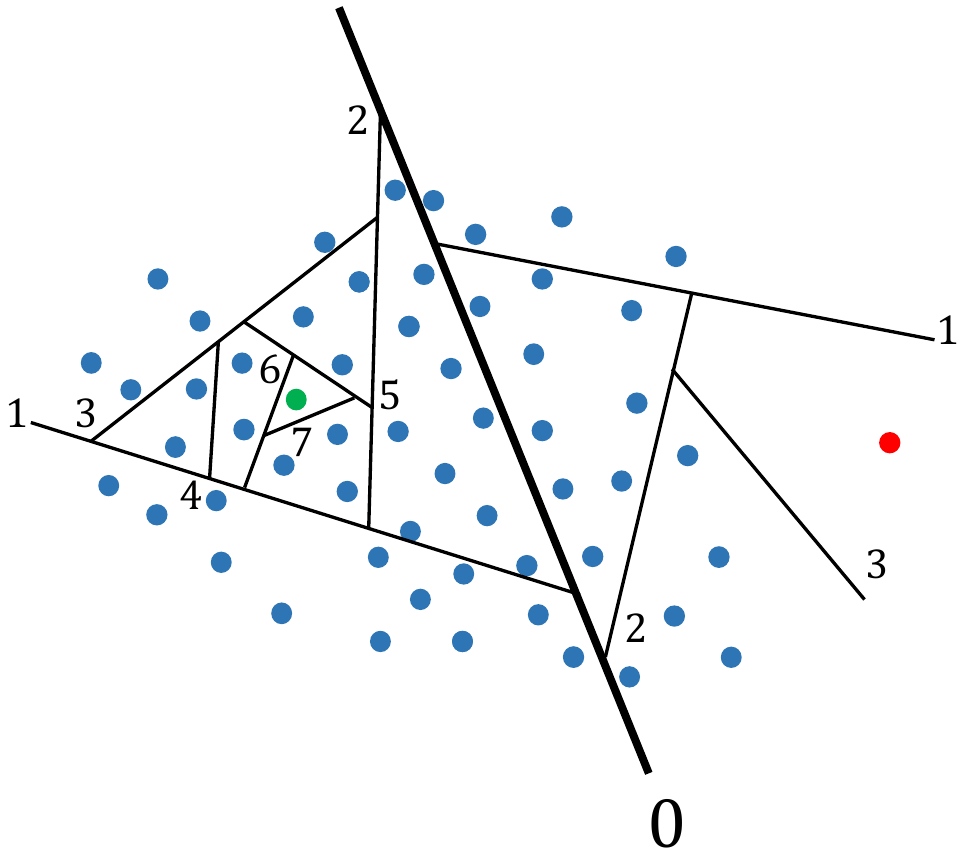}
		\label{fig3:b}
	}
	\caption{(a) When generating a new observation $I_k$, those reprojection points outside the bounding box are removed. (b) The green point needs more steps to be completely isolated than the red point, so the latter is more likely to be an outlier.}
	\label{fig3}
\end{figure}

\subsection{Outlier Removal}
The point cloud map generated by feature point-based SLAM is sparse. In the tracking thread, the feature points within bounding boxes extracted in each frame will be associated with objects. However, since objects are always located in the background rather than isolated, and are affected by occlusion and measurement noise, this process inevitably includes many outliers that do not belong to objects. We mainly use the following two methods to remove outliers.

\subsubsection{Reprojection}
Obviously, the points obtained by reprojecting the map points belonging to object onto images of different viewpoints should all be located in bounding boxes. When an object is detected in the current frame, each map point of the object is reprojected onto the image through the frame pose and intrinsic parameters. If the projection point is outside the bounding box, the corresponding map point will be removed as shown in Fig. \ref{fig3:a}. However, this method is not robust due to measurement noise, and also introduces new outliers while removing old ones.

\subsubsection{Extended Isolation Forest}
Due to continuous multi-frame observations, the points on the object surface are generally dense, while the outliers are relatively few and isolated. 
We use Extended Isolation Forest (EIF) \cite{hariri2019extended} to address this problem. Compared with the standard version, EIF uses hyperplanes with random slopes to recursively divide the sample space until each data point is isolated, while outliers require fewer steps. Fig. \ref{fig3:b} shows a demonstration.

The algorithm first constructs a binary isolation tree. For a three dimensional point cloud $\mathit{X}$, the plane normal vector $\mathbf{n}$ for each branch cut is easily obtained by sampling a standard three dimension gaussian distribution $\mathcal{N}(\mathbf{0}, \mathbf{I})$. Due to the isotropy of the standard multivariate gaussian distribution, the sampled normal vectors are uniformly distributed. Then the branching criteria for a given point $\mathbf{x} \in \mathit{X} $ is as follows:
\begin{equation}
\mathbf{x} \cdot \mathbf{n} \le p
\end{equation}
where $p$ is a random intercept that lies in the range of the branch data. If this condition is met, the point is assigned to the left branch, otherwise it is assigned to the right branch. We recursively perform the above process until each point is isolated or the depth limit is reached. By creating many such trees, we use the average depth of each point to evaluate its anomaly score:
\begin{gather}
s(\mathbf{x}, n)=2^{\frac{-E(h(\mathbf{x}))}{c(n)} } \\
c(n)=2H(n-1)-\frac{2(n-1)}{n} 
\end{gather}
where $E(h(\mathbf{x}))$ is the average depth of point $\mathbf{x}$ after traversing all trees, $c(n)$ is the normalization factor, which is only determined by the number of points $n$, and $H$ is a harmonic number. We remove those points whose anomaly scores greater than a set threshold to maintain a sparse point cloud that accurately represents object.

\subsection{Pose Estimation}

In the tracking thread, we estimate object pose using image observations and sparse point clouds. In general, objects are affected by gravity so that they are placed parallel on the support \cite{liao2022so}. We assume that the pitch and roll angles of objects are constant at zero, so only the yaw $\theta$ and translation $ \mathbf{t}  \in \mathbb{R} ^3$ need to be estimated. Object pose are represented by $T_o = \left [ R\left ( \theta \right ) \  \mathbf{t} \right ] $. First, we simply calculate the centroid of sparse point cloud to estimate $\mathbf{t}$, and then use different methods to estimate rotation according to  object appearance.

\begin{figure}[t]
	\centering
	\subfigure[]
	{
		\centering
		\includegraphics[width=1.9in]{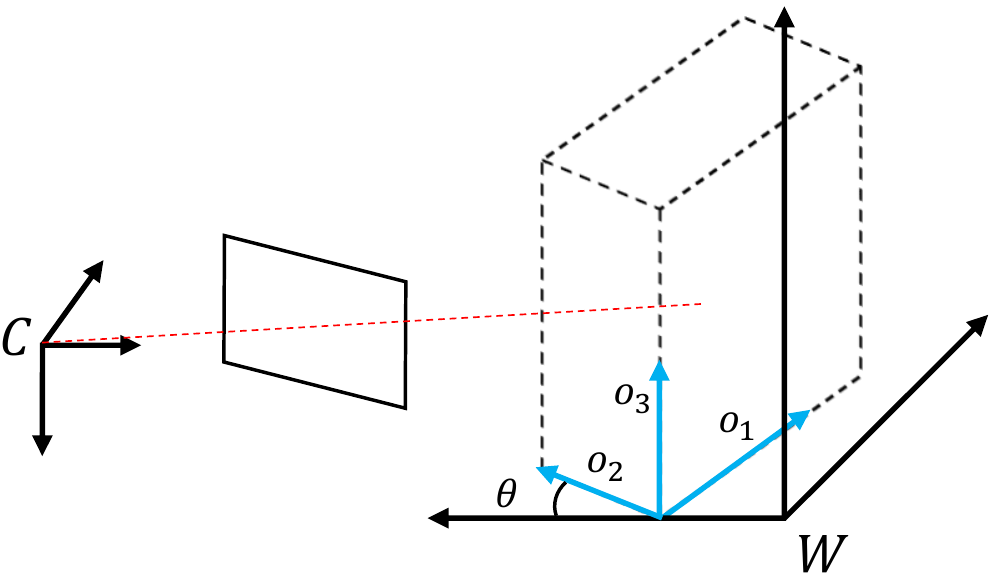}
		\label{fig4:a}
	}
	\subfigure[]
	{
		\centering
		\includegraphics[width=0.7in]{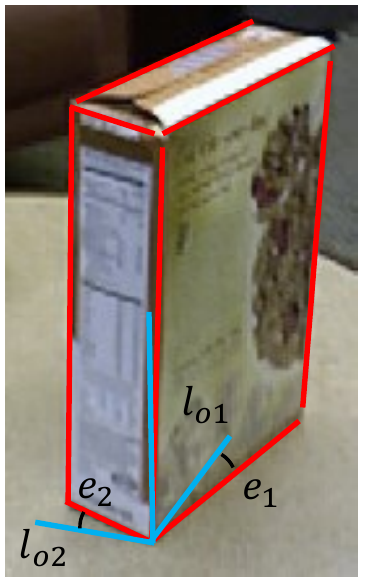}
		\label{fig4:b}
	}
	\caption{For visualization, the object frame is moved from the center to the vertices. (a) The unit line segments on the object coordinate axes are projected onto the image. (b) Optimizing the error between projected and detected line segments to estimate yaw angle.}
	\label{fig4}
\end{figure}

For objects with clear edges, such as books, chairs and keyboards, the coordinate axis of the modeled superquadric should be parallel to object edges. We first use EDLines \cite{akinlar2011edlines} to extract line segments in the image and associate those inside bounding boxes with objects. Then, as shown in Fig. \ref{fig4}, the unit line segments $l_{i}(i\in1,2,3)$ on the three coordinate axes of the superquadric are projected onto the image respectively. Finally, the accumulative angle error between the projected line segments $l_{oi}$ and the detected line segments $l_{ed}$ is optimized to estimate object yaw $\theta$. The error function is defined as follows:
\begin{gather}
\theta ^* = \mathop{\arg\min}_{\theta}\sum_{i=1}^{3} \left \| g(l_{oi})- g(l_{i:ed})\right \| ^2 \\
l_{oi} = KT_{c}^{-1} \left (  R\left ( \theta\right ) l_i + \mathbf{t}  \right ), \  i\in \left \{ 1,2,3 \right \} 
\end{gather}
where $g\left ( \cdot \right ) $ calculates the slope of line segment, and $l_{i:ed}$ is the detected line segment that matches $l_{oi}$ (angle error $<$ 5). $K$ is the camera intrinsic matrix and $T_c$ represent the camera pose.
Considering that optimizing rotation is a nonlinear process, it is crucial to have a good initial value. We uniformly sample eighteen angles from $-45^\circ $ to $45^\circ $. The sample with the largest number of matches is used as the initial value of optimization.

For objects such as bowls and cups, whose straight edges cannot be detected, we project the sparse point cloud of objects to the X-Y plane, and then directly calculate its rotation matrix corresponding to the prominent direction using principal component analysis (PCA). Please note that there is a singularity problem, i.e. rotating the superquadric coordinate frame by $90^\circ$ and swapping the scale values of two directions, the result represents the same object. We correct the PCA results by manual rotation. The angle between the $X$ axis of object frame and the $X$ axis of world frame is always less than $90^\circ$, so as to ensure that the object pose is globally consistent. Finally, we continuously update the object rotation as follows:
\begin{equation}
\theta_n = \left ( 1-\frac{1}{n}  \right ) \theta_{n-1} + \frac{1}{n} \theta_n^*
\end{equation}
where $n$ is the number of observations. With the increase of $n$, the object pose is gradually stabilized.

\subsection{Size And Shape}

After estimating superquadric pose in the tracking thread, we retrieve the remaining parameters in the local mapping thread. The sparse point cloud $P^O$ in object frame is obtained by using the object pose $T_o$. Under the influence of the robust outlier Removal algorithm, point cloud can accurately approximate the space occupation of objects, so we directly calculate the superquadric size $\mathbf{a}  = \left [ a_x,a_y,a_z  \right ] ^T$ as follows:
\begin{equation}
\mathbf{a}  = \frac{\max\left ( P^O \right )   - \min\left ( P^O \right ) }{2} 
\end{equation}
Compared to Eq. (\ref{eq2}), a better performance objective function based on radial euclidean distance is used to fit superquadric shape, which is defined as the distance from any point to the surface of the superquadric $\mathit{\Lambda}$:
\begin{equation}
G\left ( \mathit{\Lambda },\mathbf{p}^O  \right ) =\left \| \mathbf{p}^O \right \|  \left | 1-F^{-\frac{\varepsilon_1 }{2} } \left ( \mathbf{p}^O \right )  \right | 
\end{equation}
However, it is not appropriate to directly minimize the radial Euclidean distance of all points. On the one hand, some objects such as sofas and chairs are non-convex, so their sparse point clouds are not suitable for directly fitting superquadrics. On the other hand, due to the influence of observation noise and pose estimation errors, many points are not located near the surface, but are scattered inside objects. Therefore, we assign different optimization weights to errors according to the distance from the point to the center of superquadric:
\begin{equation}
\alpha_i = \frac{\left \|\mathbf{p}^O_i  \right \|-\min\left \|P^O  \right \|  }{\max\left \|P^O  \right \|-\min\left \|P^O  \right \|} 
\end{equation}
Then jointly optimize object shape, camera poses and points:
\begin{equation}
\begin{split}
\left \{ \mathit{\Lambda} , T_c,P   \right \} ^*=
\mathop{\arg\min}_{\left \{ \mathit{\Lambda},T_c,P \right \}  } &\left (  
\sum \left \| G\left (  \mathit{\Lambda}
,\mathbf{p}^O  \right ) \right \| ^2_{\Sigma_G} + \right. \\   &\left. \sum\left \| e\left (  T_c,\mathbf{p} \right )  \right \| ^2_{\Sigma_e}  \right ) 
\end{split}
\end{equation}
where $e\left ( \cdot \right ) $ is the reprojection error in traditional visual SLAM. $\Sigma$ is covariance matrix, and $\Sigma_G$ is determined by the optimization weight $\alpha$ of all shape errors. We use Levenberg-Marquardt algorithm to solve the above optimization problem. 
Similar to rotation, optimizing the above nonlinear process also requires a good initial value to avoid converging into local minima. We sample in the parameter space $\left [ 0.1,1.9 \right ]\times\left [ 0.1,1.9 \right ]$ and take the sample with the smallest distance error as the initial value.

\subsection{Data Association}

The above algorithm continuously optimizes superquadric parameters in the SLAM framework, on the premise that object observations at different times and places can be accurately associated with landmarks. Our proposed lightweight data association strategy consists of the following two parts.

\subsubsection{Inter-frame Association}

Since the bounding boxes of the same object have large overlap in consecutive images, the commonly used Intersection over Union (IoU) method is considered first. For the semantic measurement $z_i=\{b_i,c_i\}$ with bounding box $b_i$ and class label $c_i$, we traverse the object landmarks with the same label $c_i$ that were successfully associated in the previous frame, and its association is determined by the IoU of the two bounding boxes. However, this method is fragile in the case of multiple similar objects. We use inter-frame point association as a complement to the IoU method. Part of the features extracted from the same object in consecutive frames will be associated with the same map points, and this point association based on feature descriptor matching is more robust. Therefore, the shared number of the points associated with the semantic measurement and the landmark's point cloud can also be used to determine object association. We combine these two methods to improve the applicability of inter-frame association.

\subsubsection{Non inter-frame Association}
Affected by object occlusion and viewpoint changes, object detector cannot continuously output semantic measurements in each frame. Since the object centroids observed in different frames usually follow a gaussian distribution \cite{wu2020eao}, we use the single-sample t-test to deal with isolated semantic measurements in time series.
There are current measurement $z_i$ and an object landmark $O=\{\mathbf{p}_{1}^c,\dots,\mathbf{p}_{n}^c\}$, where $\mathbf{p}^c$ are the historical observations of centroid. First calculate the centroid $\mathbf{p}^c_{z_i}$ of the map points associated with $z_i$, and suppose the null hypothesis is that $z_i$ is an observation of the landmark $O$, then the $t$ statistic is defined as follows:
\begin{equation}
t = \frac{\sqrt{n}\left ( \bar{O} -\mathbf{p}^c_{z_i} \right ) }{\sigma_O  } \ \sim \   t\left ( n-1 \right ) 
\end{equation}
Given a significance level $\alpha$, the critical value from the t-distribution on $n-1$ degrees of freedom is $t_{\frac{\alpha }{2},n-1 }$. If $\left | t \right |  \le t_{\frac{\alpha }{2},n-1 }$, accept the null hypothesis and associate the semantic measure $z_i$ with the landmark $O$.

The above algorithm performs data association in real time. However, due to object detector errors or large viewpoint changes, the same object will correspond to repeated landmarks. We use the double-sample t-test to merge object landmarks. Suppose the null hypothesis is that the historical observations of two landmarks $O_1$ and $O_2$ are from the same object, then the $t$ statistic is defined as follows:
\begin{gather}
t=\frac{\bar{O}_1-\bar{O}_2 }{\sigma_p\sqrt{\frac{1}{n_1}+ \frac{1}{n_2}}  } \ \sim \   t\left ( n_1+n_2-2 \right )	\\
\sigma_p=\sqrt{\frac{\left (n_1-1  \right )\sigma_{O_1}^2 +\left (n_2-1  \right )\sigma_{O_2}^2  }{n_1+n_2-2} }
\end{gather}
where $\sigma_p$ is the pooled standard deviation. Similarly, if $\left | t \right |  \le t_{\frac{\alpha }{2},n_1+n_2-2}$, the null hypothesis is accepted. We merge the two object landmarks and re-estimate superquadric parameters.

\section{EXPERIMENTS}

To verify the performance of our proposed method, we conduct comprehensive experiments on the publicly available ICL-NUIM \cite{ICL}, TUM RGB-D \cite{TUM} and RGB-D Scenes v2 \cite{RGBDv2} datasets and compare to other state-of-the-art approaches. The 3D IoU is used to evaluate the mapping quality of objects, and it can be regarded as a comprehensive evaluation metric for object pose and shape. For the ground truth of objects, we manually label it in the global dense point cloud provided by datasets, and assign different geometric models according to the shape of objects, including cubes, ellipsoids, and cylinders. In addition, since monocular SLAM has no absolute scale, we scale the object map by aligning camera poses.

\subsection{Shape Fitting}

To verify the advantages of superquadrics as object representations, we first evaluate the quality of landmarks generated by individual objects. We select several representative objects from the results of TUM \emph{fr3\_office}, ICL-NUIM \emph{living\_room\_2}, Scenes v2 \emph{07} and \emph{09} sequences, which contain different shapes. In addition, we set the shape parameters of superquadric to $1.0$ to degenerate it into standard quadrics for comparison, both of which have the same pose and size.

\begin{figure}[t]
	\centering
	\subfigure[box]
	{
		\centering
		\includegraphics[width=0.94in]{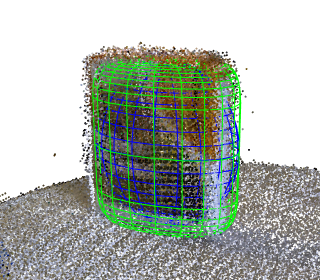}
		\label{fig5:1}
	}
	\subfigure[book]
	{
		\centering
		\includegraphics[width=0.94in]{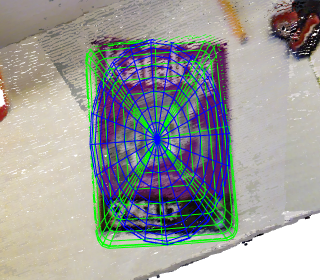}
		\label{fig5:2}
	}
\subfigure[vase]
{
	\centering
	\includegraphics[width=0.94in]{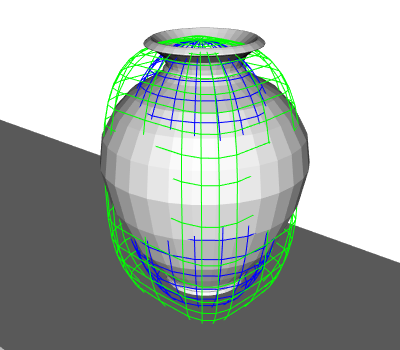}
	\label{fig5:3}
}
\subfigure[bowl]
{
	\centering
	\includegraphics[width=0.94in]{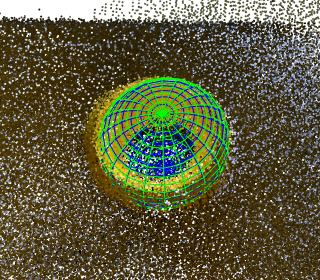}
	\label{fig5:4}
}
\subfigure[sofa]
{
	\centering
	\includegraphics[width=1.65in]{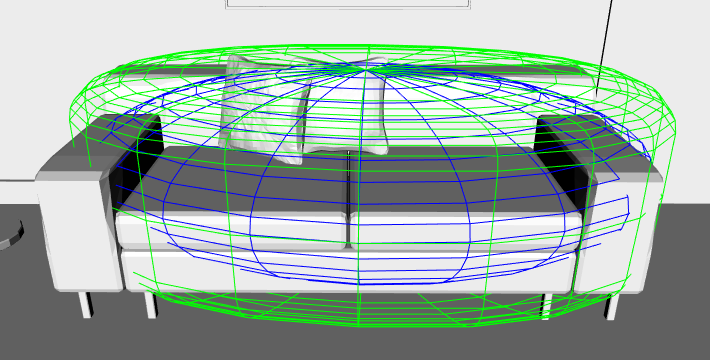}
	\label{fig5:5}
}

	\caption{Visualization of different objects, where blue and green represent quadrics and superquadrics respectively.}
	\label{fig5}
\end{figure}

The visualization results are shown in Fig. \ref{fig5}, where the object landmarks output by the system are superimposed on the dense point cloud or mesh. We can see that although the size is relatively accurate, the quadric shape is far from the actual object, and it cannot effectively represent the space occupancy information of objects. In contrast, superquadrics can more accurately fit the shape of objects, and it can adapt to different types of objects to have higher object representation strength.
Table \ref{table1} shows the quantitative results. For box, book, sofa and other similar cubic objects, superquadrics can better represent the object edges than quadrics, and the IoU is increased by $\textbf{11.9\%}$, $\textbf{9.4\%}$ and $\textbf{20.8\%}$ respectively.
For the bowl, because its surface is smooth and its shape is close to ellipsoid, the advantage of superquadrics is not obvious.

\begin{table}[t]
	\centering
	\caption{QUANTITATIVE ANALYSIS OF OBJECT REPRESENTATION}
	\label{table1}
	\renewcommand\arraystretch{0.8}
	\resizebox{!}{!}{
	\begin{tabular}{cccccc}
		
		\toprule
		Objects       & box            & book           & vase           & bowl           & sofa           \\ \midrule
		Quadrics      & 0.512          & 0.528          & 0.482          & 0.636          & 0.548          \\
		Superquadrics & \textbf{0.631} & \textbf{0.622} & \textbf{0.642} & \textbf{0.644} & \textbf{0.756} \\ \bottomrule
	\end{tabular}
	}
\end{table}

\subsection{Object Map}

\begin{figure*}[!ht]
	\centering
	\subfigure
	{
		\centering
		\includegraphics[width=1.24in]{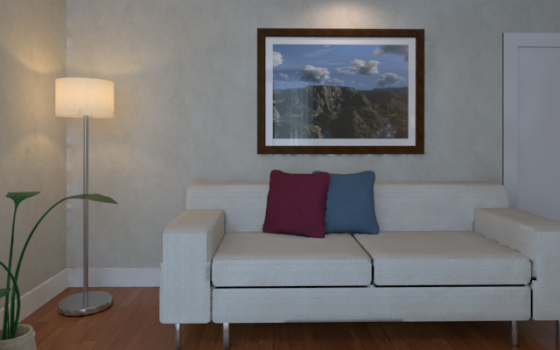}
		\label{fig6:1}
	}
	\subfigure
	{
		\centering
		\includegraphics[width=1.24in]{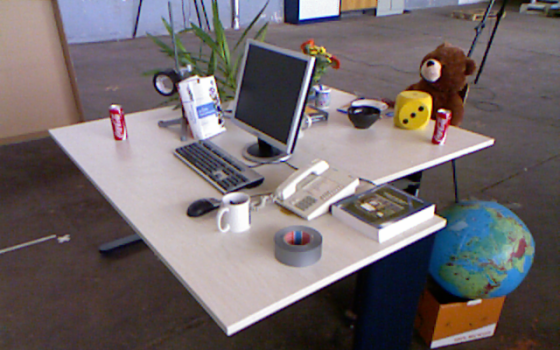}
		\label{fig6:2}
	}
	\subfigure
	{
		\centering
		\includegraphics[width=1.24in]{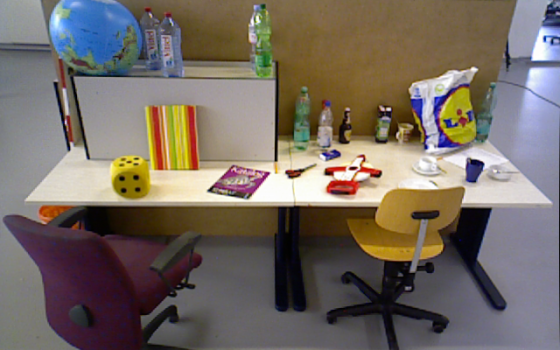}
		\label{fig6:3}
	}
	\subfigure
	{
		\centering
		\includegraphics[width=1.24in]{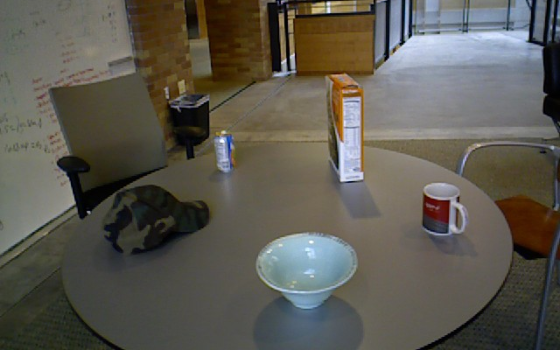}
		\label{fig6:4}
	}
	\subfigure
	{
		\centering
		\includegraphics[width=1.24in]{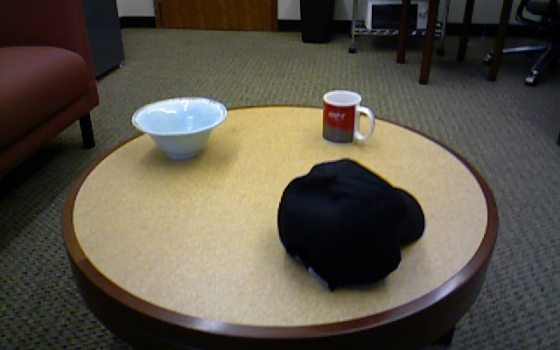}
		\label{fig6:5}
	}
	\addtocounter{subfigure}{-5}
	\subfigure[ICL\_room2]
	{
		\centering
		\includegraphics[width=1.24in]{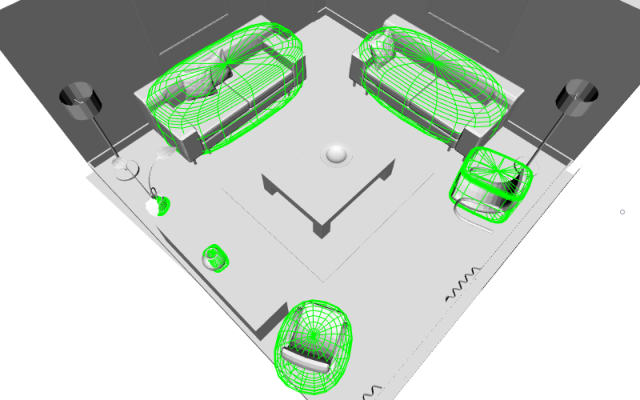}
		\label{fig6:6}
	}
	\subfigure[fr2\_desk]
	{
		\centering
		\includegraphics[width=1.24in]{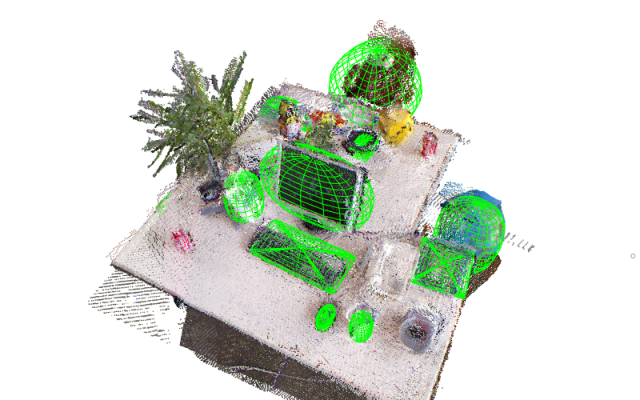}
		\label{fig6:7}
	}
	\subfigure[fr3\_office]
	{
		\centering
		\includegraphics[width=1.24in]{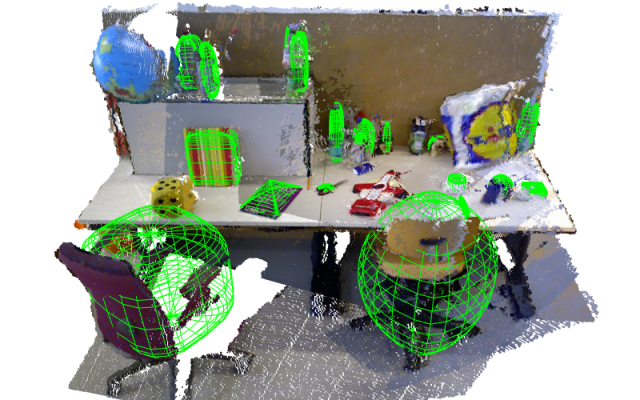}
		\label{fig6:8}
	}
	\subfigure[Sences 07]
	{
		\centering
		\includegraphics[width=1.24in]{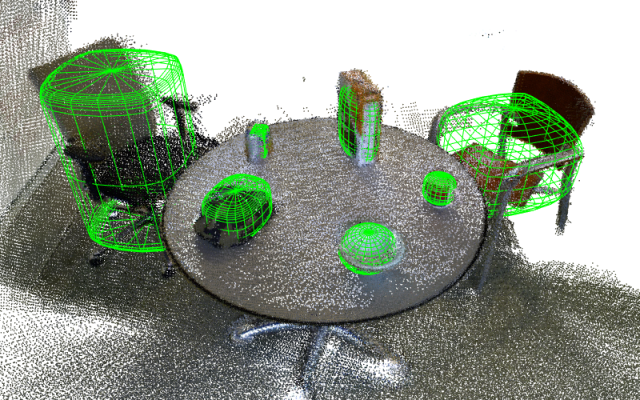}
		\label{fig6:9}
	}
	\subfigure[Sences 10]
	{
		\centering
		\includegraphics[width=1.24in]{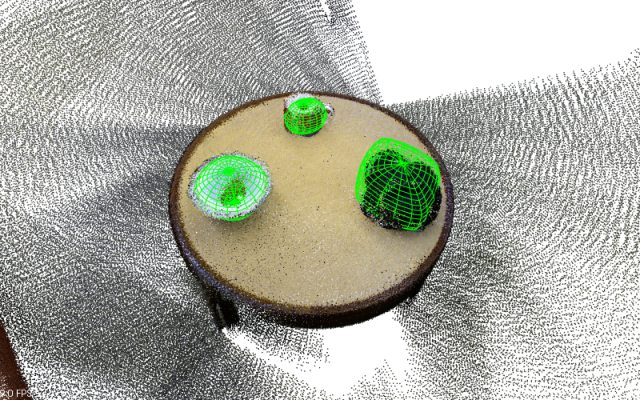}
		\label{fig6:10}
	}
	\caption{Visualization of object maps. Superquadrics can adapt to different types of objects.}
	\label{fig6}
\end{figure*}

We then evaluate the quality of the semantic object map constructed in full sequences and compare it with state-of-the-art methods QuadricSLAM \cite{nicholson2018quadricslam} and SO-SLAM \cite{liao2022so}. Unlike the automatic data association algorithm used in our system, these two methods manually associate object observations between different frames. And the number of objects evaluated is less than all objects observable in images. Therefore, we also ignore some objects with few observations or far from viewpoints. Table II presents the quantitative results. Similar to Section A, we denote the results of standard quadrics as OursQ for additional comparison, while the results of superquadrics are expressed as OursSQ.

\begin{table}[]
	\centering
	\caption{QUANTITATIVE ANALYSIS OF SEMANTIC OBJECT MAP}
	\label{table2}
	\renewcommand\arraystretch{0.8}
	\resizebox{\columnwidth}{!}{%
		\begin{tabular}{@{}cccccc@{}}
			\toprule
			& Seq         & Quadric \cite{nicholson2018quadricslam} & SO-SLAM \cite{liao2022so} & OursQ & OursSQ         \\ \midrule
			ICL-NUIM                   & room\_2       & 0.082   & 0.489   & 0.505 & \textbf{0.598} \\ \midrule
			\multirow{3}{*}{TUM}       & fr1\_desk   & 0.071   & 0.135   & 0.234 & \textbf{0.253} \\
			& fr2\_deck   & 0.172   & 0.334   & 0.411 & \textbf{0.431} \\
			& fr2\_dishes & 0.293   & 0.375   & 0.608 & 0.604          \\ \midrule
			\multirow{4}{*}{Sences v2} & 01          & -       & -       & 0.517 & \textbf{0.534} \\
			& 07          & -       & -       & 0.479 & \textbf{0.526} \\
			& 09          & -       & -       & 0.528 & \textbf{0.559} \\
			& 10          & -       & -       & 0.606 & \textbf{0.620}          \\ \bottomrule
		\end{tabular}%
	}
\end{table}

\subsubsection{ICL-NUIM dataset}

The room2 sequence records a living room scene, including large objects such as sofas and chairs. Fig. \ref{fig6:6} shows the visualization results. Our method can generate semantic object landmarks with accurate pose, and the superquadrics can fit the object shape well. Since the camera motion is gentle and there is no object occlusion, the results of oursQ are close to SO-SLAM.  In contrast, oursSQ significantly improved the landmark quality, and the IoU increased by $\textbf{10.9\%}$. This is because there are many objects in the sequence that are quite different from ellipsoids, and the superquadrics shows a great advantage.

\begin{table}[t]
	\centering
	\caption{QUANTITATIVE ANALYSIS OF DATA ASSOCIATION}
	\label{table3}
	\renewcommand\arraystretch{0.8}
	\begin{tabular}{@{}cccccc@{}}
		\toprule
		& Seq         & \cite{iqbal2018localization} & EAO-SLAM\cite{wu2020eao} & Ours       & GT \\ \midrule
		\multirow{4}{*}{TUM}       & fr1\_desk   & -       & 14              & 14         & 16 \\
		& fr2\_desk   & 11      & 22     & 22         & 26 \\
		& fr3\_office & 15      & \textbf{42}     & 38         & 45 \\
		& fr3\_teddy  & 2       & 6               & 6          & 7  \\ \midrule
		\multirow{5}{*}{Sences v2} & 01          & 5       & 7               & \textbf{8} & 8  \\
		& 07          & -       & 7               & 7          & 7  \\
		& 10          & 6       & \textbf{7}               & 6          & 7  \\
		& 13          & 3       & 3               & \textbf{4} & 4  \\
		& 14          & 4       & 5               & 5          & 6  \\ \bottomrule
	\end{tabular}
\end{table}

\subsubsection{TUM RGBD dataset}
The three sequences record different desktop scenes, in which objects are cluttered and occluded, as shown in Fig. \ref{fig6}\subref{fig6:7}\subref{fig6:8}. This is a challenge for object landmark estimation. Quadric and SO-SLAM directly use bounding box measurements to retrieve landmark parameters through the quadric projection model, which is easily affected by measurement noise and object occlusion. In contrast, our outlier removal algorithm can effectively handle object occlusion through continuous viewpoint changes and is robust to object detection noise. Quantitative results show that the quality of object maps constructed by our algorithm outperforms the comparison methods, and the superquadrics representation further improves 3D IoU. Fr2\_dishes is an exception, as it contains only ellipsoid-like objects such as bowls, plates, etc.

\subsubsection{RGBD Sences v2 dataset}
The dataset consists of 14 different tabletop scenes that contain many common small objects such as boxes, bowls, and cups. Since some objects do not belong to the scope of object detection, we evaluate on partial sequences, as shown in Fig. \ref{fig6}\subref{fig6:9}\subref{fig6:10}. Although there are textureless objects such as bowls, our algorithm can still accurately estimate pose and shape using points at the edges of objects. Similar to the previous experiments, the superquadrics representation further improves the quality of object maps, which demonstrates the effectiveness of our method.

\subsection{Data Association And Runtime Analysis}

To verify the performance of the proposed lightweight data association strategy, we compare our method with \cite{iqbal2018localization} and EAO-SLAM \cite{wu2020eao}. We use the number of object landmarks as a quantitative evaluation metric, and the results are shown in table \ref{table3}, where GT is the number of ground truth objects. For several sequences in the Sences v2 dataset, our point cloud-based non inter-frame association algorithm can easily achieve correct association due to the small number of objects and their scattered locations. The number of generated objects is very close to GT. In contrast, the TUM dataset is more challenging, such as fr3\_office shown in Fig. \ref{fig6:8}, there are many bottles placed in clusters, and these repeating objects are prone to errors during the association process. The clustering methods used by \cite{iqbal2018localization} are intractable to deal with this problem, while our method is able to correctly associate most objects and its performance is comparable to EAO-SLAM. 

Finally, we provide system runtime analysis on a laptop with an Intel Core i5-7400 CPU at 3.0 GHz and 16GB RAM. Considering that the model complexity of different detectors varies greatly, the results do not include the computational time of object detection. Table \ref{table4} shows the time usage for different numbers of objects on the TUM Fr2\_desk sequence. In general scenarios, our algorithm can run in real-time with $\textbf{25-30}$ HZ. Furthermore, our lightweight association strategy takes only $\textbf{0.7}$ ms per frame on average.

\begin{table}[]
	\centering
	\caption{AVERAGE RUNTIME OF SYSTEM}
	\label{table4}
	\renewcommand\arraystretch{0.8}
	\resizebox{\columnwidth}{!}{%
		\begin{tabular}{@{}ccccc@{}}
			\toprule
			Object number       & 0     & 4            & 8             & 12            \\ \midrule
			Tracking thread(ms) & 23.7  & 27.6 (+3.9)  & 30.4 (+6.7)   & 32.9 (+9.2)   \\
			Mapping thread(ms)  & 213.5 & 220.9 (+7.4) & 231.6 (+18.1) & 236.9 (+23.4) \\ \bottomrule
		\end{tabular}%
	}
\end{table}


\section{CONCLUSIONS}

In this work, we introduce superquadrics into SLAM to achieve accurate object representation and build semantically enhanced object maps. To do so, we propose a separate parameter estimation method and a lightweight data association strategy, both of which leverage appearance and geometric information of objects. Comprehensive experiments demonstrate the effectiveness and advantages of our method.










\bibliographystyle{IEEEtran}
\bibliography{IEEEabrv,ref}

\end{document}